\documentclass[11pt]{article}
\usepackage{nodalida2023}
\usepackage{times}
\usepackage{url}
\usepackage{latexsym}
\usepackage{graphicx}
\usepackage{url}
\usepackage{hyperref}
\usepackage{amsmath}
\usepackage{comment}
\aclfinalcopy % Uncomment this line for the final submission

\title{Student's \textit{t}-Distribution: On Measuring the Inter-Rater Reliability \\ When the Observations are Scarce}

\author{
           Serge Gladkoff$ ^{\; 1 *}$, Lifeng Han$ ^{2 *}$, % \\
\and \textbf{Goran Nenadic}$^{2}$ \\
         $^1$ Logrus Global,  Translation \& Localization  \\ 
         $^2$ The University of Manchester, UK 
         \\ {\tt lifeng.han, g.nenadic @ manchester.ac.uk} 
         \\
         {\tt %lifeng.han@adaptcentre.ie \\ \tt 
        serge.gladkoff @ logrusglobal.com} \\
        $^*$ \textit{co-first authors}}

\date{}

\begin{document}
\maketitle
\begin{abstract}
In natural language processing (NLP) we always rely on human judgement as the golden quality evaluation method. However, there has been an ongoing debate on how to better evaluate inter-rater reliability (IRR) levels for certain evaluation tasks, such as translation quality evaluation (TQE),  especially when the data samples (observations) are very scarce. In reality, practitioners need to be able to assess the reliability of human MT quality evaluation based on one, two, or maximum three human linguists' judgements.
In this work, we first introduce the little-known method to estimate the confidence interval for the measurement value when only one data (evaluation) point is available. This leads to our example with two human-generated observational scores, for which we describe ``Student's \textit{t}-Distribution'', and explain how to use it to measure the IRR score using only these two data points, and calculate the confidence interval (CI) of the quality evaluation. We give a quantitative analysis of how the evaluation confidence can be greatly improved by introducing more observations, even if only one extra observation.
We encourage practitioners and researchers to report their IRR scores and confidence intervals in all evaluations, e.g. using Student's \textit{t}-Distribution method whenever possible; thus making the NLP evaluation more meaningful, transparent, and trustworthy.

\end{abstract}

\section{Introduction}

%\textit{On human eval, confidence intervals, inter-rator reliability / agreement. }

Human evaluations have been always the gold standard to judge the quality of natural language processing (NLP) system's outputs \cite{han-etal-2021-TQA,google2021human_evaluation_TQA,gladkoff-han-2022-hope}.
This applies to many sub-tasks including machine translation (MT) \cite{han-etal-2020-alphamwe,han2022investigation,charlampidou_Gladkoff_2022NeTTT,milad2021comparative_arab},
text summarisation \cite{bhandari-etal-2020-evaluating_summarization,latif2009correlation_summary}, question answering \cite{Haifa2020etal_QA}, information extraction \cite{Han_Wu_etal2022_PLM4clinical,nenadic2004enhancing_term}, and prediction \cite{yang2009text_predict}, as well as domain applications such as social media, biomedical and clinical domains knowledge representation \cite{milovsevic2019web_summary,yang2021mining_stroke_KG,krauthammer2004term_biomedical}. 
Nonetheless, human evaluations have been subject to criticisms and debates about their reliability, particularly when conducted without strictly defined procedures. \cite{han2022overview_mte,han_gladkoff_metaeval_tutorial2022,DBLP:journals/nle/GrahamBMZ17}. 
%\cite{han2022overview_mte}
%\cite{han_gladkoff_metaeval_tutorial2022}
Despite the inclusion of factors such as quality controls and clear guidelines, human evaluation results can vary greatly among different individuals due to subjective judgements influenced by factors such as backgrounds, personalities, cultures, and so on.

Naturally, the confidence levels of human evaluation become the key point to the validity of such work.
There have been some efforts made on how to measure the confidence level of human evaluations from a statistical point of view, such as very recent work using Monte Carlo Sampling Simulations by \newcite{gladkoff-etal-2022-measuring}. 
However, this kind of statistical measurement still needs a good amount of data points, or \textit{observations}, to be based upon. 
When there are a limited amount of observations obtained from the experiments, \textit{how to measure the confidence level properly} is still a challenging question. 
One of the solutions to address this is to calculate the inter-rater agreement level and inter-rater reliability (IRR) scores. There are some historical IRR measurement metrics including Cohen's Kappa \cite{cohen1960coefficient} and Krippendorff's Alpha \cite{krippendorff1987association,krippendorff2011computing}. However, as the last issue with statistical sampling, both Cohen's Kappa and Krippendorff's Alpha need a certain amount of samples data for probability calculation, which becomes troublesome when the observations are really \textit{scarce}, e.g. only one, two, or a few data points. %three data points. 
In addition, there are existing criticisms regarding the undesired prediction of Cohen's Kappa, e.g. \newcite{delgado2019cohen_avoid_classify} gave examples about how Kappa produced better scores for worse classifiers.

In this study, we examine scenarios in which observations from human evaluations are extremely scarce (e.g., limited to one or two values) and explore potential solutions. This endeavor is motivated by the realities of the translation and localization industry. Practitioners often need to determine the reliability of human machine translation (MT) quality evaluations based on the judgments of a single linguist, or at most, two to three linguist evaluations.

We start from one observation and introduce a confidence estimation method borrowed from \newcite{abbott1962two_stage_1oberve,furnival1989confidence_1trial} which was applied to forest study by \newcite{valentine1991confidence_single_forest}. Then we discuss how this one observation-based confidence estimation is problematic and not much reliable. 
Following this, we bring an example of MT evaluation where two observations are obtained from the human assessment. 
In this case study, we introduce how to apply \textbf{Student's \textit{t}-Distribution} to measure IRR with detailed formula interpretation and guidance. We also further give instructions on how to measure confidence intervals (CIs) using this method.
We discuss the much improvement achieved by using two observational data points and Student's \textit{t}-Distribution regarding narrowed-down CIs.
Finally, we suggest that researchers also apply Student's \textit{t}-Distribution to other NLP tasks and even beyond, i.e. outside of NLP tasks. 

The rest of the paper is organised as below: Section \ref{sec_related} surveys the related work to ours on measuring IRR and confidence intervals from different fields, Section \ref{sec_single_observe} presents a case study with a single observation, Section \ref{sec_data_2plus} follows up with two and more observations where we introduce Student's \textit{t}-Distribution, and Section \ref{sec_discussion} concludes the paper with discussion and future work.

\section{Related Work}
\label{sec_related}

%\textit{Related work: on inter-rater reliability calculation, discussion, issues, on confidence intervals, our own TQE work on Monte Carlo sampling lrec22.}
%\textit{Future work, different methods, except for ‘student’ models to measure such data scarce situation? More simulations using different size of sample data.}

Regarding agreement measurement, Cohen's Kappa metric was defined by \newcite{cohen1960coefficient} as below:

\begin{comment}
\begin{align} 
\text{Penalty}&=LP0.5 \times ( \frac{\#\text{chunks}}{\#\text{matched unigrams}})^3\\
\text{MEREOR}&=\frac{10PR}{R+9P}\times (1-\text{Penalty})
\end{align}
\end{comment}

\begin{align} 
\textsc{k}&=\frac{p_o-p_e}{1-p_e}
\end{align}

\noindent  where $p_o$ is used to represent the ratio (proportion of units) when the raters agree with each other, while $p_e$ is the agreement expected by chance. In the expression of frequencies, the Kappa value can be calculated by:

\begin{align} 
\textsc{k}&=\frac{f_o-f_e}{N-f_e}
\end{align}

\noindent  In other words, the Kappa value reflects the agreement level (or proportion of agreement) after deducting the chance agreement.
In the perfect situation, when the raters all agree with each other, i.e. the $p_o$ value equals 1, the Kappa value will be 1 \cite{cantor1996sample_cohen_K}. 
However, if the raters totally disagree with each other, i.e. the value of $p_o$ is almost the same value of agreement by chance $p_e$, the Kappa value will be close or equal to 0. However, Kappa's value can be a negative number, when the agreement exhibition level is even smaller than by chance, e.g. using the above equations when the value of $p_o-p_e$ or $f_o-f_e$ is negative.
As mentioned in the earlier section, the Kappa value requires a certain number of observations to properly estimate the metric scores.

Looking into the IRR measurement in crowd-sourcing human evaluation domain, \newcite{wong-etal-2021-crossRR} argued that the traditional Krippendorff’s alpha or Cohen’s kappa threshold values, e.g. above 0.6, are not ideal due to the ignorance of cultural and individual differences from crowd-source workers. They proposed a cross-replication reliability method based on Cohen's kappa and tested the methods on human judgements of facial expressions using a large amount of 4 million data points.

%on above mentnioned in last section

%\textit{focusing on NLP / MTE/TQE, our lrec22 work}
From NLP and MT field, \newcite{Alekseeva-etal-2021-MonteCarlo,gladkoff-etal-2022-measuring} applied Monte Carlo Simulation Analysis method to generate more samples for statistically estimating the confidence intervals of judgements when the samples presented for the human evaluation are small. Their experimental outputs demonstrate that not less than 100, and ideally 200 segments are necessary for the test set to produce an unbiased, statistically significant quality evaluation of the MT system output.

\begin{comment}
    \url{https://www.researchgate.net/publication/5370070_Computing_inter-rater_reliability_and_its_variance_in_the_presence_of_high_agreement} proposed AC1, a new method

\end{comment}

Outside of NLP fields, there are also some efforts made to address similar issues in measuring reliability and confidence intervals.
For instance,
\newcite{hallgren2012computing_IRRoverview} gave a tutorial on measuring IRR for the psychology domain when multiple coders are involved using case studies using commonly used Cohen's kappa and intra-class correlation (ICC).
Similarly in  the educational and psychological domain, \newcite{walker2020using_DIF4IRR} carried out a study on applying differential item functioning (DIF) analysis to measure IRR, in comparison to the inter-class correlation coefficient and Cohen's kappa statistics.

From animal behaviour studies, \newcite{harvey2021simple_IRR_Animal} raised the issues on inter-rater and intra-rater reliability and made a discussion on Cohen's Kappa and Krippendorff's Alpha values. 
From the sociological domain, \newcite{doi:10.1177/Belur-etal2021sociological-IRR} reported a systematic survey on reporting IRR values from crime studies on multiple coders. They made further discussion on how human factors affect decision-making and how important it is to report accuracy, precision and reliability from screening/coding.

% next \url{https://bmcmededuc.biomedcentral.com/articles/10.1186/s12909-016-0788-9}?
 There is some existing business software integrating IRR into their statistical tools such as SPSS that has been used in different sectors including medical training assessment \cite{beck2016analyses_IRR_medical_train}. The IBM SPSS uses interclass correlation coefficient (ICC) to measure the IRR values among different groups of raters \footnote{\url{https://www.ibm.com/uk-en/spss}}.

%Walker CM, Göçer Şahin S. Using Differential Item Functioning to Test for Interrater Reliability in Constructed Response Items. Educ Psychol Meas. 2020 Aug;80(4):808-820. doi: 10.1177/0013164419899731. Epub 2020 Jan 20. PMID: 32616959; PMCID: PMC7307492.

However, to the best of our knowledge, there is no existing work on applying Student's \textit{t}-Distribution for measuring IRR in NLP applications, especially in translation quality evaluation (TQE) field. %NLP domain applications.

\section{On Single Judgement}
\label{sec_single_observe}
%\textit{how to measure the confidence? previous available methods, e.g. the forest research model and its theoritical support from earlier papers,  our examples in MTE.}
%\textit{leading to the next suggestion of more than two data points/ judgements }

When observational data is very scarce, more than half a century ago, 
\newcite{abbott1962two_stage_1oberve} proved the possibility of measuring confidence intervals on \textit{a single data point} from a mathematical point of view, and the later work from \newcite{furnival1989confidence_1trial} further elaborated Abbott and Rosenblatt's formula with a more narrowed interval generation. We name it the \textbf{ARF} Interval by taking the initial letters of their names. 
\footnote{
In another study by \newcite{rodriguez1996confidence_1observe} on confidence intervals (CIs) from one single observation, Herbert Robbins non-parametric CI was obtained and another technique was introduced for obtaining CI for the ``scale parameter of finite length in the logarithmic metric’’.}

This method may appear statistically counter-intuitive, but it is certainly worth mentioning here, particularly as production decisions are frequently based on a single quality measurement. An intriguing paradox arises: while many statisticians would argue it's impossible to determine a confidence interval from one measurement, project managers often rely on a single TQE (Translation Quality Evaluation) value to make their decisions. In actuality, conclusions about the reliability of a single measurement can be made, but they require supplemental information, e.g., for translation industy, known vendor's past performance. Within the ARF interval calculation method, this additional data is also derived from an experimentally-based prior knowledge or theoretically-based value that, while external to the measurement, arises from the project context. Interestingly, project managers who use a single measurement's value to make their decisions apply a similar intuition. Consequently, it's fascinating to explore what mathematical principles can elucidate within this context. 

The width of the confidence intervals reflects the uncertainty of the experiments, i.e., the wider it is, the less knowledge is available about the setup.

A relatively narrowed confidence interval indicates the controlled situations, for instance, the normal distribution in the following formula of standardised transformation:

\begin{align} 
\textsc{z}&=\frac{y - \mu}{\sigma}
\end{align}

\noindent of which, $y$, $\mu$, and $\sigma$ are the variables of the \textit{variate}, \textit{mean value}, and \textit{standard deviation}. The parameter $z$ represents the \textit{standardised variate}. 

For the situation with one observation, let $\hat\mu$ be the independent and fixed value that is known before and outside of the measurement, and $y$ be the experimental measurement value. \newcite{furnival1989confidence_1trial} gives the following calculation intervals:

\begin{align} 
ARF = \frac{y + \hat{\mu}}{2}\pm k|y-\hat{\mu}|
\end{align}

\noindent This ARF interval contains the probability of $\mu$ that is larger than or equal to $1-\alpha$, and $\alpha$ meets the following equation with $k$:

\begin{align} 
k = \frac{1-\alpha + \sqrt{1-2\alpha}}{2\alpha} \; , \; 0 < \alpha \leq 0.5
\end{align}

The pair value of ($k$, $\alpha$) was given by \newcite{furnival1989confidence_1trial} as in Table \ref{tab:k_alpha_match}.

\begin{table*}[!t]
    \label{hope_metric}
    \centering
    \begin{tabular}{cccccccc}
    \hline
    & &&&  value ($\alpha$)&&& \\ \hline
    \textbf{Distribution} & 0.50 & 1/3 &0.25 & 0.2 & 0.1& 0.05 & 0.01  \\ \hline
    Normal ($k$) & 0.05 &	1.26&	1.8&	2.31	&4.79	&9.66	&48.39	 \\
    Unknown ($k$) &0.5  &1.87  &2.91 &3.94 &8.97  &18.99 &99 
\\    \hline
\end{tabular}
\caption{The Value Matching of ($k$, $\alpha$) for both Normal Distribution and Unknown Ones by \newcite{furnival1989confidence_1trial}.} \label{tab:k_alpha_match}
\end{table*}

\subsection{Case Study using ARF Intervals}
Let's have a case study on using ARF intervals for MT evaluation. Assuming that a translation vendor has been evaluated earlier on certain lines of projects and the average result was a score of 96.3 ($\hat\mu$). 
The next translation quality measurement produced by another vendor is a lower rating of 85.2 ($y$). How reliable is this measurement by itself purely from the statistical point of view? and what does it tell us?
If we assume the quality measurements are distributed normally, it is logical to take the average value of the prior history evaluations as the predicted value for future outcomes. Below we give two practices using ARF intervals.

1) If we construct a 75\% confidence interval for the true quality rating, we need to use k=1.8 from the instruction by Table \ref{tab:k_alpha_match}, and the corresponding $\alpha$ value is 0.25. Using the ARF interval formula, it gives:

\begin{align} 
ARF = \frac{96.3 + 85.2}{2} \pm 1.8 \times |96.3-85.2|
\end{align}

\noindent which is 90.75$\pm$19.98. Therefore, the ARF interval for the true value of quality rating is [70.77, 100].
As we can see from this example, the 75\% confidence interval is almost half of the measured value itself, i.e. the maximum deviation of 19.98 is 22\% of the measurement result (90.75). Although the mathematical precision of the single quality measurement is limited to this level, it can be beneficial to define these limits.

2) Similarly, for an 80\% CI ($\alpha = 0.2$) for the true quality rating, the corresponding $k$ value from the Table \ref{tab:k_alpha_match} is 2.31 and the above formula gives the following ARF value:

\begin{align} 
ARF = \frac{96.3 + 85.2}{2} \pm 2.31 \times |96.3-85.2|
\end{align}

\noindent which is 90.75$\pm$25.64. In other words, the interval for the true value of quality rating is [65.1, 100]. From this, we can see that with 80\% CI, the maximum deviation of 25.64 is 25\% of the measurement result (90.75). 

From these two case studies, what is probably more interesting in the context of translation quality evaluation is that ``the middle of the CI lies halfway between an earlier average result and the recent lower measurement''. We can spell a good rule of thumb:  \textit{if the single measurement deviates from the average, the true value is likely halfway between the average and the new measurement}.
Knowledge of this would help not to overreact to unusually low single scores newly generated.

Even though it is possible to measure the confidence levels, this ARF interval is very wide and the worse thing is that it can not be improved by the choice of $\alpha$.
As shown in Table \ref{tab:k_alpha_match}, 0.5 is the narrowest option of choice for intervals. However, this value is considered not high enough to make a significant impact. 
To the right side of the table, the smaller value the $\alpha$ is, the wider the resulting intervals will be. 
Therefore, choosing $\alpha$ value between (0.2, 0.25) is probably the compromise to make when there is only one observation or judgement available. 
From this case study, our finding is that evaluations consisting of only a single measurement are not recommended as there will be a higher chance of bias as illustrated by our translation evaluation example.
Such measurements have only rough and indicative values, so the data collection and analysis approaches must be invoked to improve the quality of measurement itself with the data science apparatus.
This will lead to our next section when we recommend that a second quality measurement is very necessary, how to measure it in the new situation, and how much difference it will make.

\section{On Observations of More Than One}
\label{sec_data_2plus}
Following the last section, we \textbf{call on more measurement points} for NLP evaluation tasks, especially in the language service provider sector where the single observation value is still very common in practice due to the cost concern. \footnote{e.g. referring to the R\&D report from Language Service Provider \url{https://logrusglobal.com/}}

For instance, when a single translation quality measurement is not satisfactory for one of the parties, second quality measurement can be made to validate the first measurement.
Then, how much improvement to the confidence interval can be obtained by introducing extra observational data?
To answer this, 
the obvious problem, of course, is that at least 20-30 data points are required to calculate the mathematical variance for a normal distribution.
In settings where the sample size is less than 30, and the standard deviation of the entire population is unknown, \textit{Student’s t-Distribution} can be used to evaluate standard deviation based only on the number of measurements between one and 30, e.g. 2, 3, etc.

%...serge: ``The obvious problem, of course, is that at least 20-30 data points are required to calculate the mathematical variance for a normal distribution. In settings where sample size is less than 30, and the standard deviation of the entire population is unknown, Student’s distribution can be used to evaluate standard deviation based only on the number of measurements between one and 30 (2, 3, etc. measurements).''

\subsection{On Student's \textit{t}-Distribution}

When the sample size (\textit{aka} observations) is very small in comparison to the entire population, \newcite{student1908probable} designed 
Student's \textit{t}-distribution to measure the mean errors and the confidence intervals of estimation.

\begin{figure*}[!t]
\begin{center}
\centering
\includegraphics*[width=0.90\textwidth]{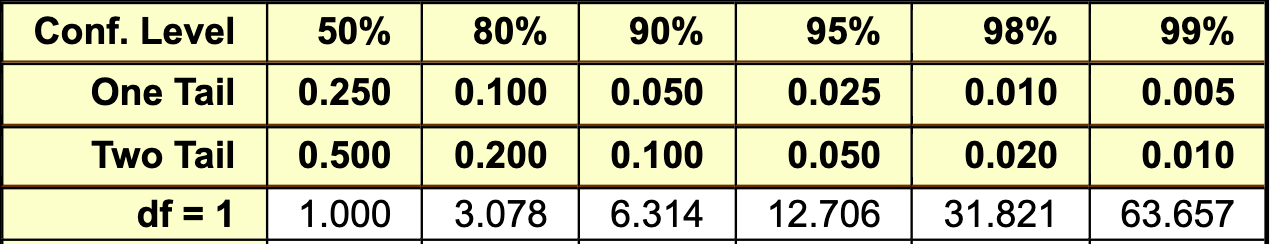}
\caption{Critical Values for Student's \textit{t}-Distribution with One Degree of Freedom (from people.richland.edu) }
\label{fig:critical-t-value-1free}
\end{center}
\end{figure*}

When there is one degree of freedom, the critical values for Student's \textit{t}-Distribution are shown in Figure \ref{fig:critical-t-value-1free} including the confidence level, one tail, and two tail scores. \footnote{\url{https://people.richland.edu/james/lecture/m170/tbl-t.html}} 
The full list of critical values with more degrees of freedom is shown in 
Figure \ref{fig:critical-t-value-list}.\footnote{\url{https://www.stat.purdue.edu/~lfindsen/stat503/t-Dist.pdf}} There are many researchers who proposed different algorithms to calculate these critical values by hand and using computers, for instance, the work from \newcite{cheng1983algorithm_t_x_f} and comparison studies by \newcite{blair1980comparison_t_rank}.

% https://courses.lumenlearning.com/introstats1/chapter/a-single-population-mean-using-the-student-t-distribution/#:~:text=Concept%20Review&text=The%20t%2Dscore%20follows%20the,n%20is%20the%20sample%20size.
The notation of Student's \textit{t}-Distribution is defined as below if we use \textit{T} as the random variable:

\begin{itemize}
    \item  $T \sim t_{df}$ where $df=n-1$
\end{itemize}
where $df$ is the degree of freedom and $n$ is the number of observations. For instance, if we have a sample size $n=2$, we calculate the $df=2-1=1$ and write the distribution as $T \sim t_{1}$.

For the situation when the standard deviation is unknown, the error bound for the sample mean is defined as:

\begin{align} 
E =  (t_{\alpha/2})(\frac{s}{\sqrt{n}})
\end{align}

\noindent where $t_{\alpha/2}$ is the critical value of t-score with the area to the right equal to $\alpha/2$ (Figure \ref{fig:critical-t-value-list}), $s$ is the standard deviation of observations (samples): % https://www.khanacademy.org/math/statistics-probability/summarizing-quantitative-data/variance-standard-deviation-sample/a/population-and-sample-standard-deviation-review

\begin{align} 
s = \sqrt{\frac{\sum(x_i-\overline{\rm x})^2}{n-1}}
\end{align}
where $\overline{\rm x} $ is the mean value of $n$ samples:

\begin{align} 
\overline{\rm x} = (1/n)\sum{x_i}	
\end{align}

The resulting confidence interval (CI) is then the following span:

\begin{align} 
CI = (\overline{\rm x} - E, \overline{\rm x} + E)	
\end{align}

%\begin{figure*}[!t]
% https://people.richland.edu/james/lecture/m170/tbl-t.html

\subsection{Deploying \textit{t}-Distribution to IRR}

Looking back to our MT evaluation experiments, from a practical industry project on language service, we have an example to demonstrate how to deploy Student's \textit{t}-Distribution to measure IRR value.
Assume we have used the Multidimensional Quality Metric (MQM) initialised by \newcite{MQM2014} \footnote{open source project \url{https://themqm.org/}} and professional translators for a translation evaluation project and got two numbers of overall quality scores: QS1=76.85 and QS2=81.99 on a scale from 0 to 100.\footnote{This is a real example from an industrial project on TQE called ``Whale''.} 
We can immediately see that the QS2 81.99 is 6.7\% greater than the QS1 76.85, and oppositely QS1 76.85 is 6.3\% less than QS2 81.99. 
Therefore QS2 agrees with 93.3\% of QS1, and QS1 agrees with 93.7\% of QS2. 
This is almost 95\% agreement, so it looks good for most cases. 
However, if the PASS/FAIL threshold is 80, the difference may be crucial for the translator.  Then, \textit{what is the reliability of this evaluation result}?

The Sample Mean $\overline{\rm x} $  of QS1 and QS2 is $(QS1+QS2)/2=79.42$ for this sample of two measurements.
The Sample Standard Deviation $s$ for this sample of two values is $\sqrt{(QS1-\overline{\rm x})^2+(QS2-\overline{\rm x})^2}=\sqrt{6.6049\times2}$ which is 3.6345.

The Confidence Interval depends on the desired Confidence Level, which, in turn, depends on the subject matter area of the content which was translated.
For most fields, the confidence level should be at least 80\%. The critical number $t_{\alpha/2}$  for that level (0.1 which is 20\% divided by 2 for one tail of the graph, $\alpha=0.2$) and two measurements (one degree of freedom, \textit{df}=1) is 3.078, as shown in Figure \ref{fig:critical-t-value-1free} and \ref{fig:critical-t-value-list}. 

Therefore, the margin of error for these measurements is:

\begin{align} 
E =  \frac{3.078\times 3.6345}{\sqrt{2}}=7.91
\end{align}

%E = 12.71 * 2.57 / SQRT (2) = 5.6
\noindent This means that the confidence interval for these two measurements is 7.91$\times$2=15.8, which indicates that with an 80\% degree of confidence, the true quality score lies on this interval:
[79.42 – 7.91, 79.42 + 7.91], or [71.51, 87.33].

As we can see from this result, given the second measurement, we can significantly improve the confidence interval as compared to the single measurement. 
The two different judgements (observations) that we obtained reduce variance from 25\% of a single judgement (Section \ref{sec_single_observe}) to 9.96\% (7.91/79.42), i.e. more than two times narrower with an 80\% confidence setting. %22\% from a single judgement to 7\%, i.e. 5.6/79.42.

However, as in the previous example, this confidence interval is still relatively large. Can we tell anything about the translator passing or failing the 89\% PASS/FAIL threshold? The answer is that since the sample mean is below 80\% and equals 79.42, the evaluation result is borderline FAIL.

Ideally, we need a third measurement or even more observations to further improve this interval, but %the good news is that this is already a fairly narrow interval for such a subtle, almost intangible factor as the human perception of quality.
in a production setting, the additional data points are rarely obtained by repeated evaluations, due to the cost and time constraints required for such a process. 

The good news is that we already have reliable information for translation quality evaluation (TQE) purposes: 
this is a borderline FAIL, becasue the Sample Mean is lower than the threshold, and therefore more than half of possible values are below the PASS threshold. 
This is not bad for the measurement of such a subtle, almost intangible object as the human perception of quality.
But you can only obtain a history of performance based on multiple evaluations for different content pieces and apply data science approaches.

\begin{comment}
    \footnote{\url{https://www.stat.purdue.edu/~lfindsen/stat503/t-Dist.pdf}}. below. 
More on NIST web \footnote{\url{https://www.itl.nist.gov/div898/handbook/eda/section3/eda3664.htm}} and \footnote{\url{https://www.itl.nist.gov/div898/handbook/eda/section3/eda3672.htm}}

\end{comment}
% https://mathworld.wolfram.com/Studentst-Distribution.html
%\cite{student1908probable}

\section{Conclusion and Future Work}
\label{sec_discussion}

When it comes to evaluating translation quality, the ability to measure alone is not sufficient; we also need to know how reliable the measurement is. Automatic evaluation of quality quickly produces the same scores if repeated a number of times, which creates an illusion of precision. Unfortunately, the results of automatic quality measurement not only depend on the language pair, the Machine Learning system, the decoder, and the content type, but also vary from dataset to dataset, depending on the way the data have been cleaned and formatted. Given these factors, automatic measurement can be very fast and “reliable”, but it may be (and often is) invalid, as well as inconsistent.
Human translation quality evaluation (TQE) is currently the only way to obtain valid measurements of human perception of quality, and considered to be the golden standard of TQE. However, human measurement’s inter-rater reliability (IRR) should be assessed, even if evaluation has been carried out correctly. Even if evaluators are experienced linguists, trained to do evaluation according to proper system, and client specifications are clearly defined, the evaluators would still produce close but not the same evaluation results due to the very nature of human perception of quality, which is by definition the function of personal perception.
This problem is exacerbated by the fact that in real life production setting there is no time or budget to validate the translation quality measurement even with the second reviewer, and even if there is a second reviewer, the low IRR of such measurement makes it difficult to confirm the first measurement.

In this paper, we first studied the typical production setting of gold standard human quality measurement, where TQE is performed by only one experienced, trained linguist, according to clearly defined customer specifications, producing a single measurement, and make conclusions about the reliability of such measurement. 
We then illustrated the results with the case of\textit{ Student's \textit{t}-Distribution analysis of two measurements} made by two different reviewers.

From the first and second experiments, we can conclude that a single measurement has very low reliability and only has an indicative value. 
The confidence interval for one measurement is (as shown in Section \ref{sec_single_observe}) as wide as 25\%, and therefore one evaluation cannot be taken as a basis for process decisions, more measurements are required. For instance, the second measurement can narrow down this interval and render it two times smaller, to around 10\% (Section \ref{sec_data_2plus}).

Yet we can say that the middle of the confidence interval lies halfway between the earlier average result and the lower recent measurement. \textit{A good rule of thumb is born:}
\textbf{if the single measurement deviates from the average, the true value sits, in all probability, halfway between the average and the new measurement}. 
Consequently, the recommendation is: please do not over-react on an unusually low new single score, take a middle ground between the older average and the new score, and think about it as the most probable result. 

The second measurement may improve the confidence interval significantly but is rarely done unless during the arbitration. 
Therefore, it is more practical to obtain additional data points from other evaluations of different samples, in the course of the translation process.

Subsequent evaluations effectively are placed into two categories: (a) mostly PASS with only rare occasional FAIL, (b) all other cases (mostly FAIL, or many FAILs). This strategy is caused by the desire to ensure that a system is reliably well above the PASS/FAIL threshold and thus ensures quality results. 
Multiple evaluations also confirm the validity of quality measurements and allow the application of well-known maths of statistics of normal distribution.

However, it is worth noting that proper methods of data analysis are required to analyse \textit{quality evaluation data-sets}, such as:

\begin{itemize}
    \item Removal of outliers, which are caused by irrelevant causes.
    \item Evaluations made on very small or very large samples.
    \item Evaluations that are incorrect due to the improper application of metrics such as counting repeated errors, for example.
    \item Evaluations made by reviewers who were not trained, subjective, or had in mind different customer requirements.
\end{itemize}

It should be remembered, that data science only allows obtaining good results if you clean the data properly. Incorrect, biased, not properly calibrated, or imprecise conclusions and inferences may result from using uncleaned data.

\section*{Limitations}
In this work, we discussed how to calculate confidence intervals and evaluation reliability when there are only one or two assessment scores from annotators, such as  translation quality assessors. For the first case when there is only one new observation score, we assume there is a pre-estimated/expected score ready to use, i.e. for ARF interval. However, this might not be the case in some situations, or it might cost some time and money to get this value, for instance, for a newly established task without much prior knowledge. 
In the second case considered, we introduced Student's \textit{t}-distribution method and gave two human judgement scores for estimation. This is expected to be helpful for the small number of observations; however, it does require some mathematical calculations using guided formulas and parameter tables, which might be not instantly convenient to translators or project managers who do not have much statistical knowledge, and requires manual calculations from educated AI researchers anyway. For real world applications preliminary setup and additional clear and crisp guidance for practitioners may be required.

\section*{Ethical Statement}
There are no ethical concerns in this work since it is only about introducing alternative methodologies for calculating the confidence and reliability of human evaluations.

\section*{Acknowledgements}
 We thank the Whale project for sharing their real TQE examples with us to deploy the discussed methods.
 We thank Dr. Leonore Findsen at Purdue University Statistics Department and Prof. James Jones of Mathematics at Richland Community College for their open-source lecture notes.
 We are grateful to the anonymous reviewers for their valuable comments on our work.
This work has been partially supported by grant EP/V047949/1 ``Integrating hospital outpatient letters into the healthcare data space'' (funder:
UKRI/EPSRC). %∗ The authors LH and GN thank the project support from HIPS (R126035 task A05) and from JigSaw (R124782 task A07) at The University of Manchester.
%\section{}

\bibliographystyle{acl_natbib}
\bibliography{nodalida2023}

\section*{Appendix}
\begin{figure*}[!t]
\begin{center}
\centering
\includegraphics*[width=0.99\textwidth]{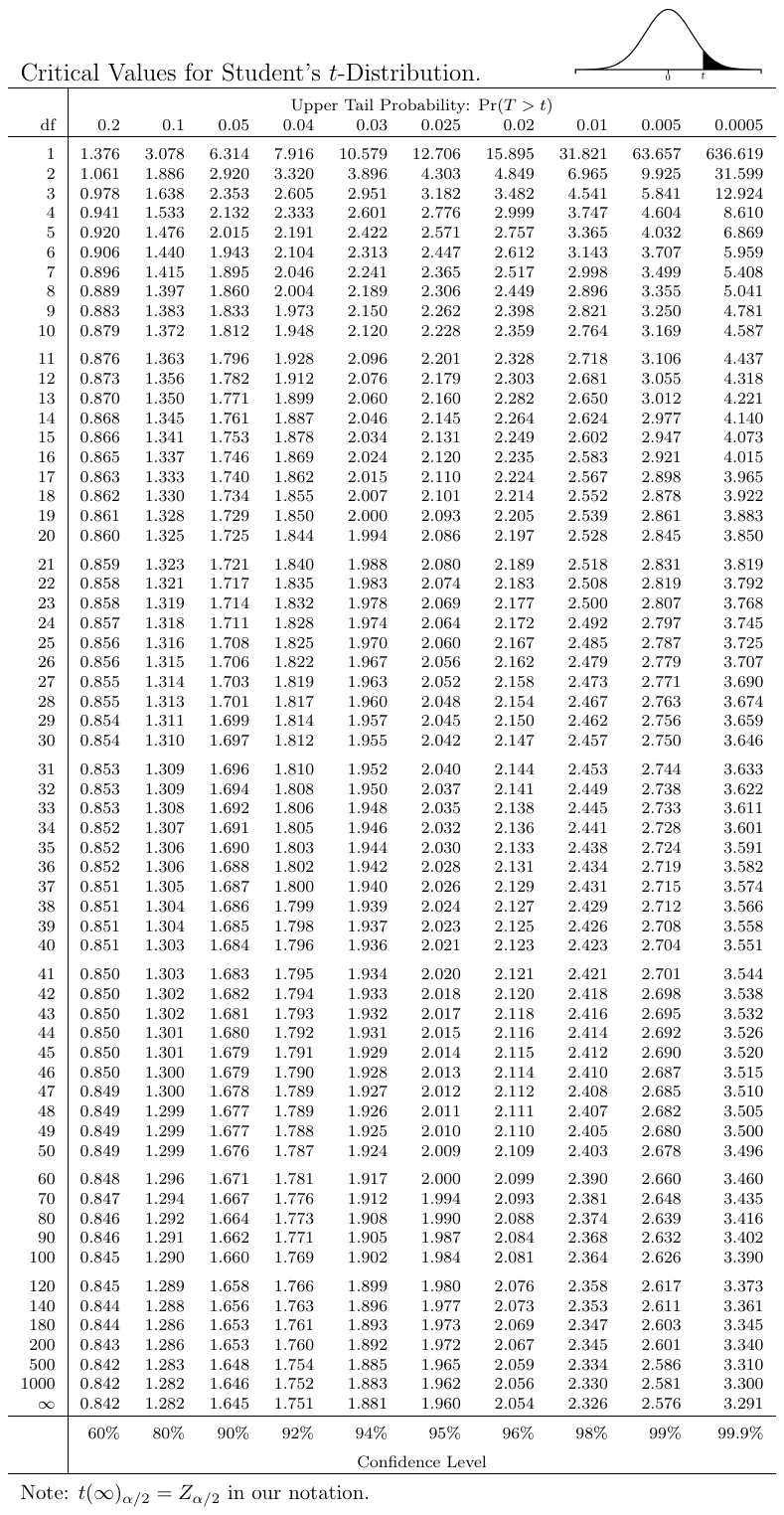}
\caption{Critical Values List for Student's \textit{t}-Distribution (from stat.purdue.edu) }
\label{fig:critical-t-value-list}
\end{center}
\end{figure*}

A detailed Critical Value from the Student's \textit{t}-Distribution is displayed in Figure \ref{fig:critical-t-value-list} from Purdue University Statistics.
%Figure \ref{fig:critical-t-value-list}
%\begin{figure*}[!t]
% t-Dist-stat-purdue-edu.pdf 

\end{document}